\documentclass[10pt,journal,compsoc]{IEEEtran}
\usepackage{cite}
\usepackage{url}
\usepackage{ragged2e}
\usepackage{footnote}
\usepackage{animate}
\makesavenoteenv{tabular}
\makesavenoteenv{table}
\usepackage{graphicx}
\usepackage{amsmath,amssymb} 
\usepackage{mathtools} %
\usepackage{algorithm}
\usepackage{algorithmicx}
\usepackage{algpseudocode}
\usepackage{graphics}
\usepackage{multirow}
\usepackage{booktabs}
\usepackage{subcaption}
\usepackage{pifont}%
\usepackage[pagebackref=true,breaklinks=true,colorlinks,citecolor=cyan,linkcolor=red,bookmarks=false]{hyperref}

\newcommand{\myPara}[1]{\noindent\textbf{#1}}
\newcommand{\cmark}{\ding{51}}%
\newcommand{\xmark}{\ding{55}}%

\newcommand{\secref}[1]{Sec.~\ref{#1}}
\newcommand{\tabref}[1]{Table.~\ref{#1}}
\newcommand{\figref}[1]{Fig.~\ref{#1}}
\renewcommand{\eqref}[1]{Eq.~\ref{#1}}
\newcommand{\ie}{\textit{i.e.}}
\newcommand{\eg}{\textit{e.g.}}

\usepackage{makecell}

\begin{document}

\title{Long-Context Autoregressive Video Modeling with Next-Frame Prediction}

\author{
  Yuchao Gu, Weijia Mao, Mike Zheng Shou
  \IEEEcompsocitemizethanks{
      \IEEEcompsocthanksitem Corresponding author: Mike Zheng Shou (E-mail: mikeshou@nus.edu.sg)
    \IEEEcompsocthanksitem Yuchao Gu, Weijia Mao and Mike Zheng Shou are with the Showlab, National University of Singapore.
    }
}

\IEEEtitleabstractindextext{%
\begin{abstract}
  \justifying
Long-context video modeling is essential for enabling generative models to function as world simulators, as they must maintain temporal coherence over extended time spans.
However, most existing models are trained on short clips, limiting their ability to capture long-range dependencies, even with test-time extrapolation. While training directly on long videos is a natural solution, the rapid growth of vision tokens makes it computationally prohibitive.
To support exploring efficient long-context video modeling, we first establish a strong autoregressive baseline called \textbf{\textit{\underline{F}}}rame \textbf{\textit{\underline{A}}}uto\textbf{\textit{\underline{R}}}egressive (FAR). FAR models temporal dependencies between continuous frames, converges faster than video diffusion transformers, and outperforms token-level autoregressive models.
Based on this baseline, we observe \textbf{\textit{context redundancy}} in video autoregression. Nearby frames are critical for maintaining temporal consistency, whereas distant frames primarily serve as context memory. To eliminate this redundancy, we propose the long short-term context modeling using \textbf{\textit{asymmetric patchify kernels}}, which apply large kernels to distant frames to reduce redundant tokens, and standard kernels to local frames to preserve fine-grained detail. This significantly reduces the training cost of long videos.
Our method achieves state-of-the-art results on both short and long video generation, providing an effective baseline for long-context autoregressive video modeling. The code is released at \url{https://github.com/showlab/FAR}.
\end{abstract}

}

\maketitle

\IEEEdisplaynontitleabstractindextext

\IEEEpeerreviewmaketitle

\section{Introduction}
\label{sec:intro}
\IEEEPARstart{L}ong-context video modeling is essential for advancing video generative models toward real-world simulation~\cite{videoworldsimulators2024}. However, current state-of-the-art video generative models (\eg, Wan~\cite{wang2025wan}, Cosmos~\cite{agarwal2025cosmos}) fall short in this aspect. These models are typically trained on short video clips; for example, Wan and Cosmos are trained on approximately 5-second video segments. These methods effectively capture short-term temporal consistency, such as object or human motion. However, they fail to maintain long-term consistency, like memorizing the observed environments, which requires learning from long video observations.

Learning from long videos presents inherent challenges, primarily due to the prohibitive computational cost associated with processing the large number of vision tokens. As a result, many previous efforts have focused on test-time long video generation, employing training-free approaches to produce extended video sequences~\cite{lu2024freelong,zhao2025riflex}. However, while these methods can generate visually plausible long videos, they fail to effectively leverage long-range context.
To truly capture long-range dependencies, it is essential to train or fine-tune models directly on long videos. However, this remains computationally expensive. Concurrent efforts, such as direct long-context tuning~\cite{guo2025long}, face high computational costs, while test-time training~\cite{dalal2025one} typically requires specialized architectural designs and incurs additional inference overhead. Therefore, an efficient framework for long-video training is urgently needed to enable effective long-context video modeling.

\begin{figure}[!tb]
    \centering
\includegraphics[width=\linewidth]{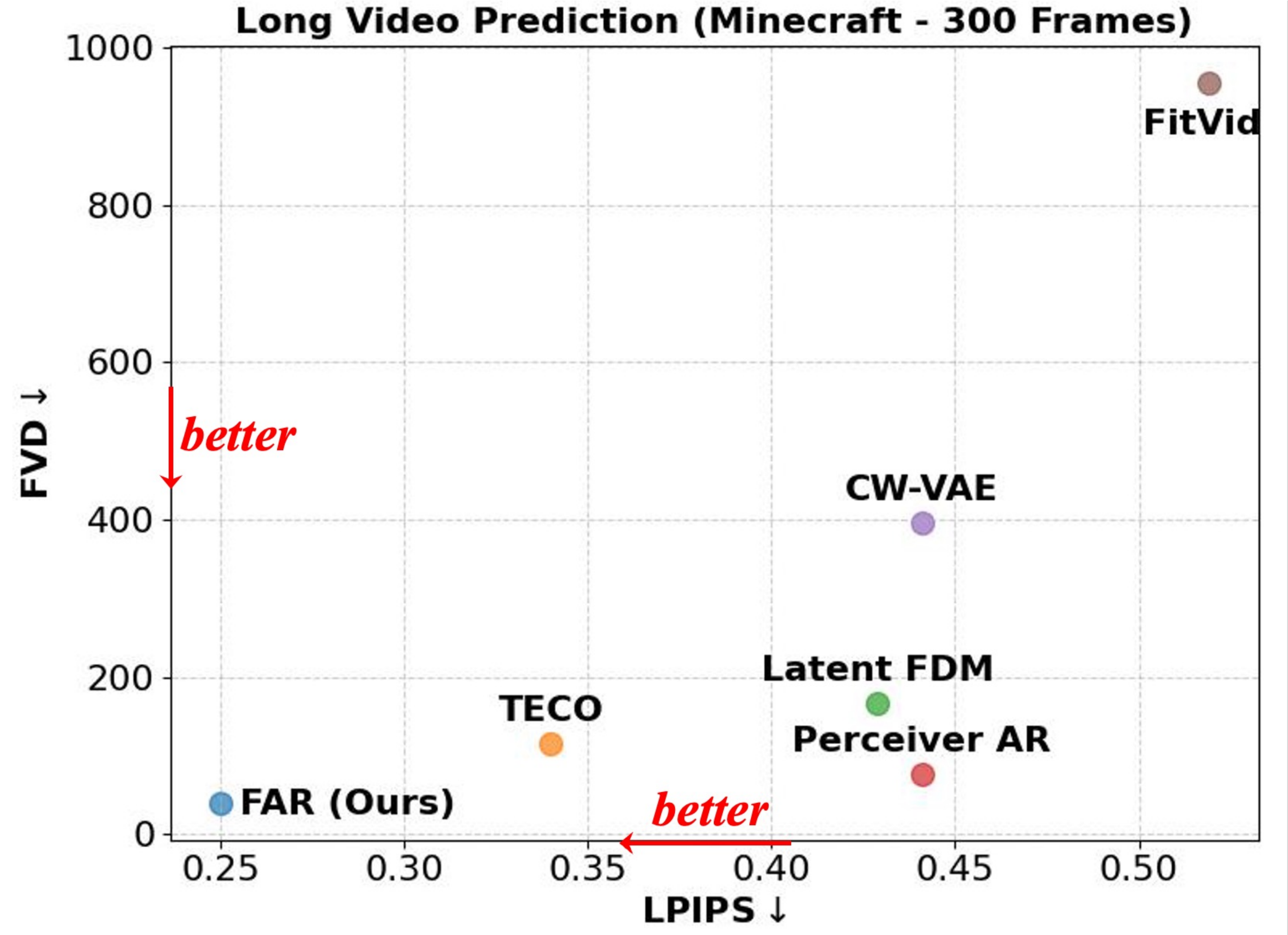}
    \vspace{-.25in}
    \caption{\textbf{Evaluation on Long Video Prediction.} FAR better exploits long video contexts and achieves accurate prediction.}
    \label{fig:teaser}
    \vspace{-.1in}
\end{figure}

To investigate this problem, we first establish a baseline for video autoregressive modeling, namely the \textbf{\textit{\underline{F}}}rame \textbf{\textit{\underline{A}}}uto\textbf{\textit{\underline{R}}}egressive (FAR) model. Unlike token-based autoregressive models (Token-AR), FAR operates in a continuous latent space, capturing causal dependencies between frames while still allowing full attention modeling within each frame. FAR is trained using a frame-wise flow matching objective with autoregressive context.
As a hybrid AR-Diffusion model, FAR also encounters a gap between the observed context during training and inference. This is a common issue in similar models (\eg, ~\cite{chen2025diffusion}, \cite{jin2024pyramidal}, \cite{xie2024show}).
To address this problem, we propose training FAR with \textbf{\textit{stochastic clean context}}, which enables the model to leverage clean context signals during training and thus reduces bias at inference time.
We demonstrate that FAR trained with stochastic clean context achieves better performance than video diffusion transformers and Token-AR, establishing itself as a strong autoregressive video generation baseline.

Building on FAR, we observe \textbf{\textit{context redundancy}} in video autoregressive modeling. Specifically, the current frame relies more heavily on nearby frames to capture local motion consistency, while distant frames primarily function as context memory. To exploit this property, we introduce long short-term context modeling with \textbf{\textit{asymmetric patchify kernels}}. In this approach, we apply a large patchify kernel to distant context frames in order to compress redundant tokens, while using the standard patchify kernel on nearby context frames to preserve fine-grained temporal consistency. This method significantly reduces training costs on long videos and leads to notable improvements in FAR’s long-context modeling capability, as demonstrated in action-conditioned long video prediction.

Our contributions are summarized as follows:
\begin{enumerate}
\item We introduce FAR, an strong autoregressive video generation baseline, combined with stochastic clean context to bridge the training-inference gap in observed context. 
\item Building on FAR, we observe context redundancy in video autoregressive modeling and propose long short-term context modeling with asymmetric patchify kernels to substantially reduce long-video training costs.
\item FAR achieves state-of-the-art performance in both short- and long-video modeling.
\end{enumerate}

\section{Related Work}

\subsection{Video Generation}
\myPara{Video Diffusion Models.}
Recent advances in video generation have led to the scaling of video diffusion transformers~\cite{yang2024cogvideox,videoworldsimulators2024,kong2024hunyuanvideo} for text-to-video generation, resulting in superior visual quality.
Pretrained text-to-video models are subsequently fine-tuned to incorporate images as conditions for image-to-video generation~\cite{guo2023animatediff,xing2024dynamicrafter,yang2024cogvideox}.
The trained image-to-video models can be utilized for autoregressive long-video generation using a sliding window~\cite{lu2024freelong,wang2023gen}, but their ability to leverage visual context is limited by the sliding window's size.
In this work, we show that FAR achieves better convergence than video diffusion transformers for short-video generation while naturally supporting variable-length visual context.

\myPara{Token Autoregressive Models.}
Video generation based on token autoregressive models (\ie, Token AR) aims to follow the successful paradigm of large language models. These models typically quantize continuous frames into discrete tokens~\cite{yu2023language,gu2024rethinking} and learn the causal dependencies between tokens using language models~\cite{kondratyuk2023videopoet,hong2022cogvideo}.
While they achieve plausible performance, their generation quality remains inferior to that of video diffusion transformers due to information loss from vector quantization. Additionally, unidirectional visual token modeling may be suboptimal~\cite{fan2024fluid}.
Subsequent studies have explored continuous tokens~\cite{li2025autoregressive} without vector quantization but have not demonstrated their effectiveness in video generation. In this work, we show that FAR can learn causal dependencies from continuous frames and achieve better performance than Token AR in both short- and long-video modeling.

\myPara{Hybrid AR-Diffusion Models.}
To leverage the strengths of both continuous latent spaces and autoregressive modeling, recent studies~\cite{xie2024show, zhou2024transfusion, ma2024janusflow} have explored hybrid AR-Diffusion models. These models typically employ a diffusion objective for image-level modeling with autoregressive contexts. Hybrid AR-Diffusion models are widely applicable to both visual~\cite{xie2024show, jin2024pyramidal, chen2025diffusion} and language generation~\cite{barrault2024large, wu2023ar}. Recent research has also applied it in frame-level autoregressive modeling~\cite{jin2024pyramidal, chen2025diffusion} for video generation. However, they suffer from a training-inference discrepancy in the observed context. Some studies~\cite{hu2024acdit, zhou2025taming} have attempted to mitigate this issue by maintaining a clean copy of the noised sequence during training, but this approach doubles the training cost.
Among these methods, FAR efficiently addresses the training-inference gap through the proposed stochastic clean context, demonstrating its superior performance in long-context video modeling.

\begin{figure*}[!tb]
    \centering
\includegraphics[width=\linewidth]{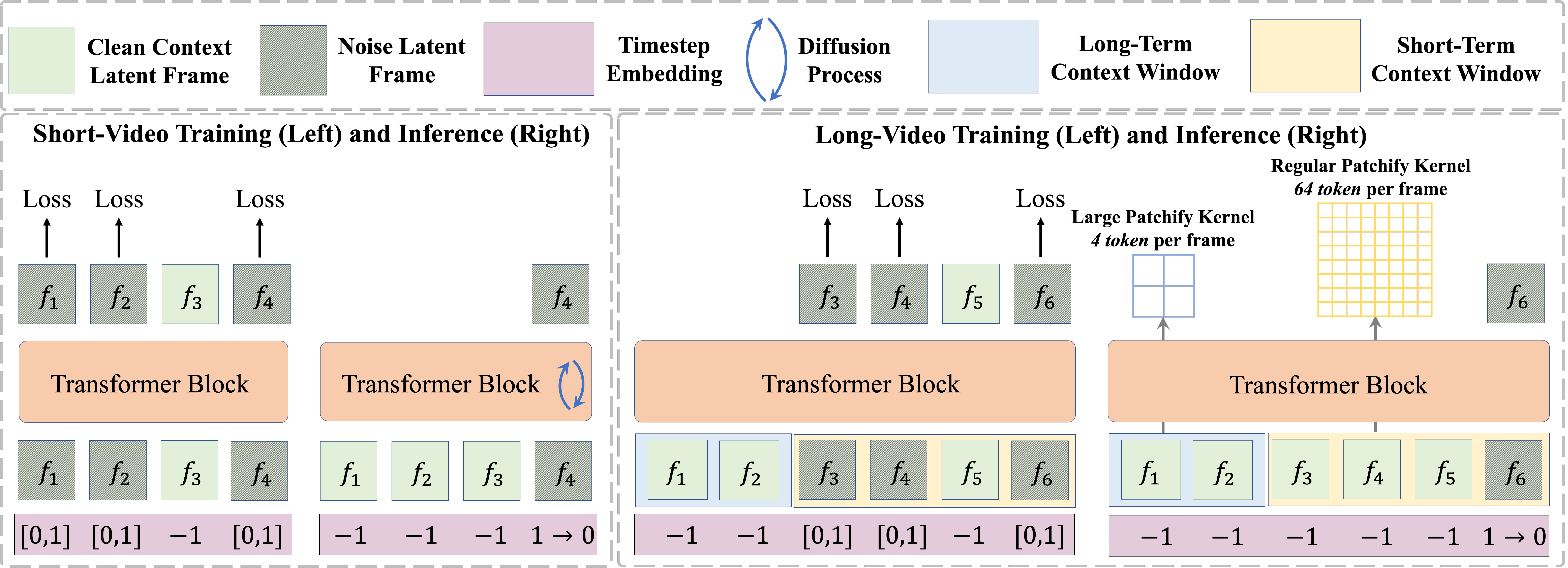}
    \vspace{-.27in}
    \caption{\textbf{Illustration of FAR's Training and Inference Pipeline.} In short-video training, a portion of frames is randomly replaced with clean context frames, marked with a unique timestep embedding (\eg, -1) beyond the flow-matching scheduler. In long-video training, we adopt long short-term context modeling. A long-term context window with aggressive patchification is adopted to reduce redundant vision tokens, while a short-term context window is used to model fine-grained temporal consistency.}
\label{fig:pipeline}
\vspace{-.1in}
\end{figure*}

\subsection{Long-Context Language Modeling}
Long-context language modeling typically follows two main approaches: test-time extrapolation and direct fine-tuning on long sequences. In the test-time extrapolation setting, many studies explore extrapolatable positional embeddings~\cite{press2021train,peng2023yarn,blocntkaware}. While these methods enable inference on longer sequences, they often underperform compared to models trained directly on long-text corpora. To reduce the computational cost, recent work~\cite{chen2023extending, chen2023longlora} has proposed efficient  long-sequence fine-tuning strategies for large language models. However, training on long video sequences poses greater challenges, as vision tokens grow much faster than language tokens with increasing context length. To address this, we propose a long short-term context modeling approach using asymmetric patchify kernels, which effectively reduce context redundancy during long-video training.

\subsection{Long-Context Video Modeling}
Recent advancements in video generation models have enabled their use as interactive world simulators~\cite{valevski2024diffusion,bruce2024genie,parkerholder2024genie2}, which require the ability to exploit long-range context and memorize the observed environment. However, existing video diffusion transformers lack effective mechanism to utilize long-range context.
Although early work~\cite{yan2023temporally,hawthorne2022general,harvey2022flexible} has explored long-video prediction, it has been limited in visual quality and long-range consistency. In this work, we introduce FAR, a efficient framework for long-context autoregressive video modeling.

\begin{table}[!tb]
    \centering
    \caption{\textbf{Model Variants of FAR.} We follow the model size configurations of DiT~\cite{peebles2023scalable} and SiT~\cite{ma2024sit}.}
    \label{tab:model_cfg}
    \vspace{-.1in}
    \resizebox{\linewidth}{!}{\begin{tabular}{lcccccc}
        \toprule
        \textbf{Models} & \textbf{\#Layers} & \textbf{Hidden Size} & \textbf{MLP} & \textbf{\#Heads} & \textbf{Params} \\
        \midrule
        FAR-B  & 12  & 768  & 3072 & 12 & 130M \\
        FAR-M  & 12  & 1024  & 4096 & 16 & 230M \\
        FAR-L  & 24  & 1024 & 4096 & 16 & 457M \\
        FAR-XL & 28  & 1152 & 4608 & 18 & 674M \\\midrule
        FAR-B-Long  & 12  & 768  & 3072 & 12 & 158M \\
        FAR-M-Long  & 12  & 1024  & 4096 & 16 & 280M \\
        \bottomrule
    \end{tabular}}
\end{table}

\section{Preliminary}
\subsection{Flow Matching}
Flow Matching~\cite{liu2022flow, lipman2022flow, albergo2022building} is a simple alternative objective for training diffusion models. Rather than modeling the reverse process with stochastic differential equations, Flow Matching learns a continuous vector field that deterministically conntect two distribution.

\begin{figure}[!tb]
    \centering
    \includegraphics[width=\linewidth]{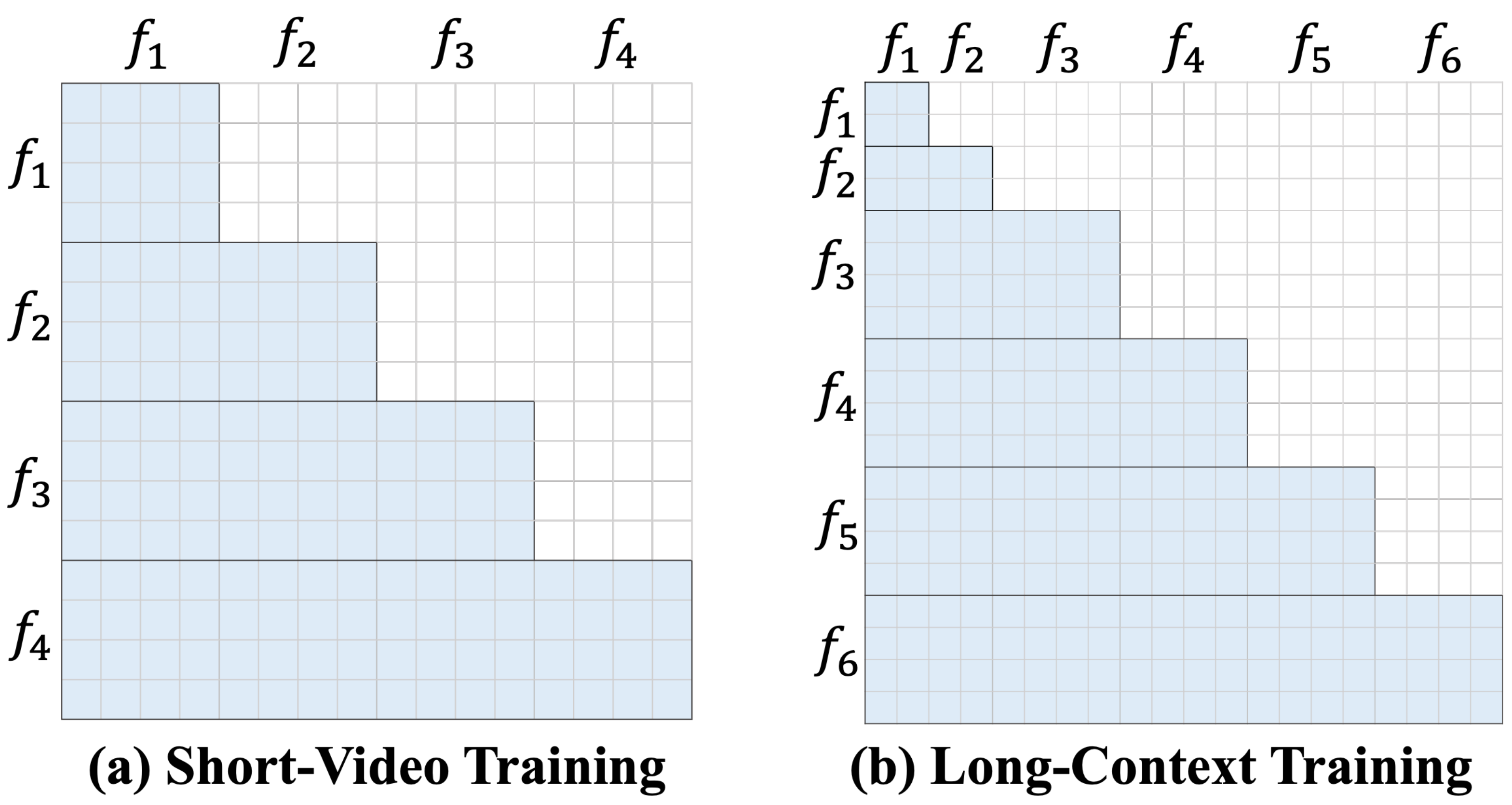}
    \vspace{-.24in}
    \caption{\textbf{Visualization of Attention Mask.} FAR enables full attention within a frame while maintaining causality at the frame level. In long-context training, we adopt aggressive patchification for long-term context frames to reduce tokens.}
    \label{fig:attention_mask}
    \vspace{-.1in}
\end{figure}

Specifically, given a data sample \( x_0 \sim p_{\text{data}}(x) \) and a noise sample \( x_1 \sim \mathcal{N}(0,I) \), we construct a continuous trajectory connecting them via linear interpolation:
{
\begin{equation}
    \label{eq:noise_input}
    x(t) = (1-t)x_0 + t x_1, \quad t \in [0,1].
\end{equation}
}This formulation implies a constant velocity:
{
\begin{equation}
    \frac{dx(t)}{dt} = v^* = x_1 - x_0.
\end{equation}
}To enable the model to learn the optimal transport between the data and noise distributions, we introduce a learnable time-dependent velocity field \( v_\theta(x,t) \). During training, a random time \( t \sim U(0,1) \) is sampled, and the model is optimized by minimizing the following objective:
{
\begin{equation}
    \label{eq:fm}
    \mathcal{L}(\theta) = \mathbb{E}_{x_0, x_1, t} \left[ \left\| v_\theta(x(t), t) - v^* \right\|^2 \right].
\end{equation}
}

\subsection{Autoregressive Models}
Autoregressive models are a class of probabilistic models where each element in a sequence is conditioned on its preceding elements, denote as context. 
Formally, given a sequence of tokens \( (x_1, x_2, \dots, x_n) \), an autoregressive model assumes that each token \( x_i \) is generated based on its previous tokens \( (x_1, x_2, \dots, x_{i-1}) \). The generative process can be expressed as a factorization of the joint probability:
{
\begin{equation}
    p(x_1, x_2, \dots, x_n) = \prod_{i=1}^{n} p(x_i | x_1, x_2, \dots, x_{i-1}).
\end{equation}}By modeling each token conditioned on its preceding tokens, autoregressive models naturally capture the sequential dependencies inherent in data.

\begin{figure}[!tb]
    \centering
    \includegraphics[width=\linewidth]{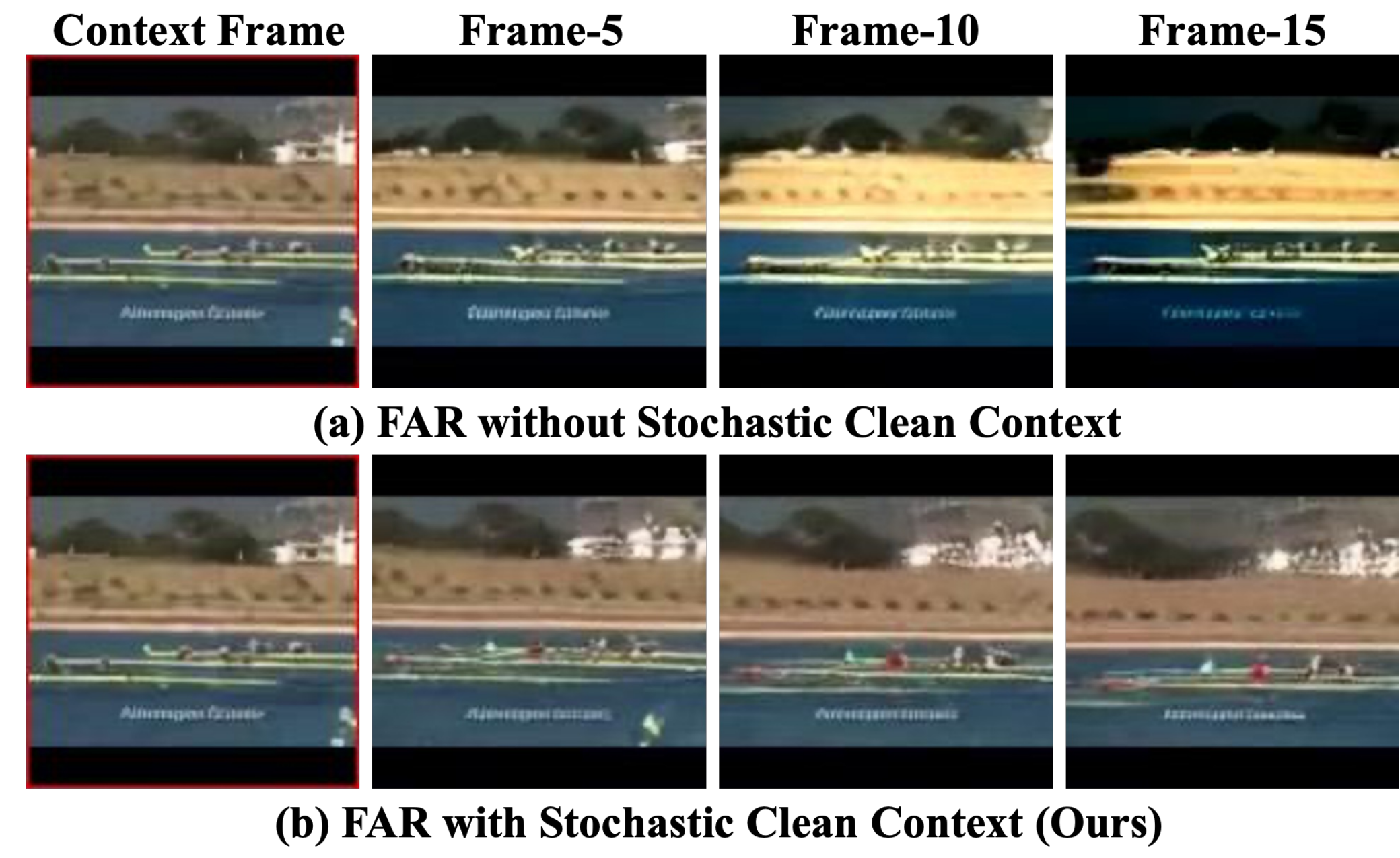}
    \vspace{-.24in}
    \caption{\textbf{Effect of Stochastic Clean Context.} This technique eliminate training-inference gap in observed context.}
    \label{fig:stochastic_clean_context}
    \vspace{-.1in}
\end{figure}

\section{FAR}

We first present the FAR framework in \secref{sec:framework}, followed by training challenges and solutions in \secref{sec:basictrain}. \secref{sec:long_context} analyzes the design for long-context video modeling, and \secref{sec:kv_cache} introduces the KV cache for faster inference.

\subsection{Framework Overview}
\label{sec:framework}

\myPara{Architecture.}
As shown in \figref{fig:pipeline} (a), FAR is built upon the diffusion transformer~\cite{ma2024sit,peebles2023scalable}. We adopt the model configuration of DiT~\cite{peebles2023scalable} and Latte~\cite{ma2024latte}, as listed in \tabref{tab:model_cfg}.
The key architectural difference between FAR and video diffusion transformers (\eg, Latte~\cite{ma2024latte}) lies in the attention mechanism. As shown in \figref{fig:attention_mask}(a), for each frame, we apply causal attention at the frame level while maintaining full attention within each frame. We adopt this causal spatiotemporal attention for all layers, instead of the interleaved spatial and temporal attention used in Latte.
In FAR, image generation and image-conditioned video generation are jointly learned thanks to the causal mask, whereas video diffusion transformer~\cite{ma2024latte} requires additional image-video co-training.

\myPara{Basic Training Pipeline.} The training pipeline of FAR is illustrated in \figref{fig:pipeline} (a). Given a video sequence $\mathbf{X}$, we first employ a pretrained VAE to compress it into the latent space $\mathbf{Z} \in \mathbb{R}^{T \times H \times W}$, where $T$, $H$, and $W$ denote the number of frames, height, and width of the latent features, respectively.
Note that although we primarily adopt an image VAE in this work, FAR can also be trained with a video VAE since our autoregressive unit is the latent frame. Following diffusion forcing~\cite{chen2025diffusion}, we independently sample a timestep for each frame. We then interpolate between the clean latent and the sampled noise using Eq.~\eqref{eq:noise_input} and apply the frame-wise flow matching objective in Eq.~\eqref{eq:fm} for learning.
The key difference between FAR and image flow matching lies in that we adopt causal spatiotemporal attention, allowing each frame to access previous context frames during denoising.

\begin{figure}[!tb]
    \centering
    \includegraphics[width=\linewidth]{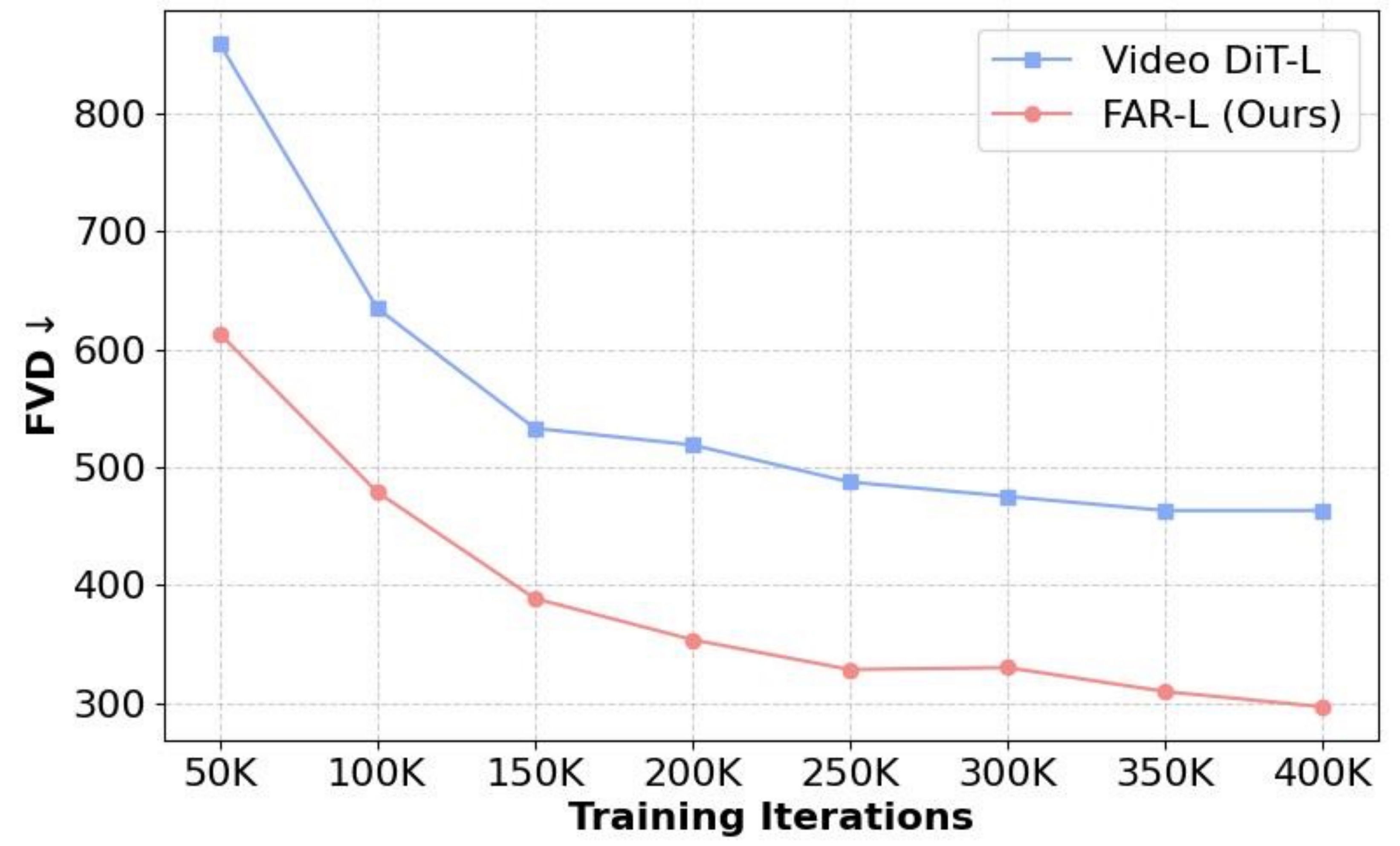}
    \vspace{-.24in}
    \caption{\textbf{Comparison of FAR and video diffusion transformer.} FAR achieves better convergence than video diffusion transformer in unconditional video generation on UCF-101.}
    \label{fig:comp_video_dit}
    \vspace{-.1in}
\end{figure}

\begin{table}[!tb]
    \centering
    \caption{\textbf{Comparison of Routes for Long-Context Video Modeling.} In test-time extrapolation, FAR is trained on short videos and evaluated on long videos using different extrapolation methods.}
    \label{tab:comp_tte}
    \vspace{-.1in}
    \resizebox{\linewidth}{!}{
        \begin{tabular}{l|cccc}
            \toprule
            \textbf{Method} & 
            \textbf{SSIM$\uparrow$} & 
            \textbf{PSNR$\uparrow$} & 
            \textbf{LPIPS$\downarrow$} & 
            \textbf{FVD$\downarrow$} \\
            \midrule\midrule
            \multicolumn{5}{c}{\textbf{Test-Time Extrapolation}} \\
            \midrule
            Sliding Window & 0.365 & 12.3 & 0.415 & 161 \\
            Naive RoPE Ext. & 0.372 & 12.2 & 0.397 & 396 \\
            RIFLEx~\cite{zhao2025riflex} & 0.372 & 12.2 & 0.398 & 391 \\
            \midrule
            \multicolumn{5}{c}{\textbf{Long-Video Training}} \\
            \midrule
            FAR-B-Long & \textbf{0.576} & \textbf{19.3} & \textbf{0.153} & \textbf{34}  \\
            \bottomrule
        \end{tabular}
    }
\end{table}

\begin{figure*}[!tb]
    \centering
    \includegraphics[width=\linewidth]{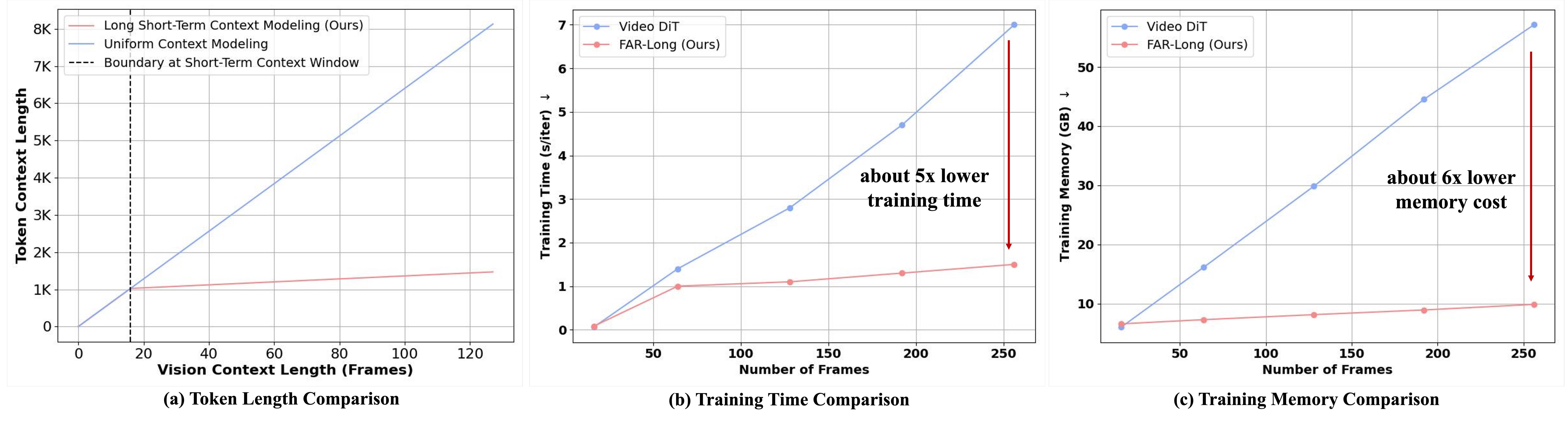}
    \vspace{-.24in}
    \caption{\textbf{Relation between Token Context Length and Vision Context Length.} With the proposed long short-term context modeling, the token context length scales more slowly with increasing vision context length compared to uniform context modeling. When training on long videos, the reduced number of tokens leads to significantly lower training costs and memory usage.}
    \label{fig:asymetric_context}
    \vspace{-.1in}
\end{figure*}

\subsection{Short-Video Modeling}
\label{sec:basictrain}
\myPara{Training-Inference Gap in Observed Context.} As a hybrid AR-diffusion model, FAR also encounters a training-inference gap in the observed context. As illustrated in \figref{fig:pipeline}(a), each clean latent is fused with sampled noise for the flow matching objective, as defined in Eq.~\eqref{eq:noise_input}. Consequently, later frames can only access the noised version of previous frames during training. However, during inference, this leads to a distribution shift when clean context frames is used.

As shown in the example in \figref{fig:stochastic_clean_context}(a), the training-inference gap in the observed context leads to a distribution shift when inferring with a clean context. Although adding mild noise to the context during inference can help mitigate this effect, it still causes low-level flickering, degrading the quality of the generated video.
Recent works~\cite{hu2024acdit, zhou2025taming} attempt to address this issue by maintaining a clean copy of the noised sequence during training. However, this approach doubles the training costs.

\myPara{Our Solution: Stochastic Clean Context.} To bridge the gap in observed context, we introduce stochastic clean context for training FAR. As illustrated in \figref{fig:pipeline}(a), we randomly replace a portion of the noised frames with their corresponding clean context and assign them a unique timestep embedding (\eg, -1) beyond the flow-matching timestep scheduler. These clean context frames are excluded from loss computation and are implicitly learned through later frames that use them as context. During inference, this unique timestep embedding guides the model to use clean context effectively. Training FAR with stochastic clean context does not add extra computation and does not conflict with different timestep sampling strategies during training (\eg, logit-normal sampling~\cite{esser2024scaling}). It effectively resolves the training-inference discrepancy, as exemplified in \figref{fig:stochastic_clean_context}(b).

\myPara{FAR \textit{vs.} Video Diffusion Transformer.} 
FAR and video diffusion transformer differ only in their training schemes. FAR is trained with independent noise and causal attention, while the video diffusion transformer is trained with uniform noise and full attention. This raises an interesting question: \textbf{\textit{Can FAR surpass video diffusion transformers?}}
To explore this, we convert FAR to video diffusion transformer as a baseline, denoted as Video DiT. We align the training settings to compare the two paradigms. As shown in \figref{fig:comp_video_dit}, FAR achieves better convergence than the Video DiT, demonstrating its potential to become a strong baseline for autoregressive video modeling.

\begin{figure}[!tb]
    \centering
    \includegraphics[width=\linewidth]{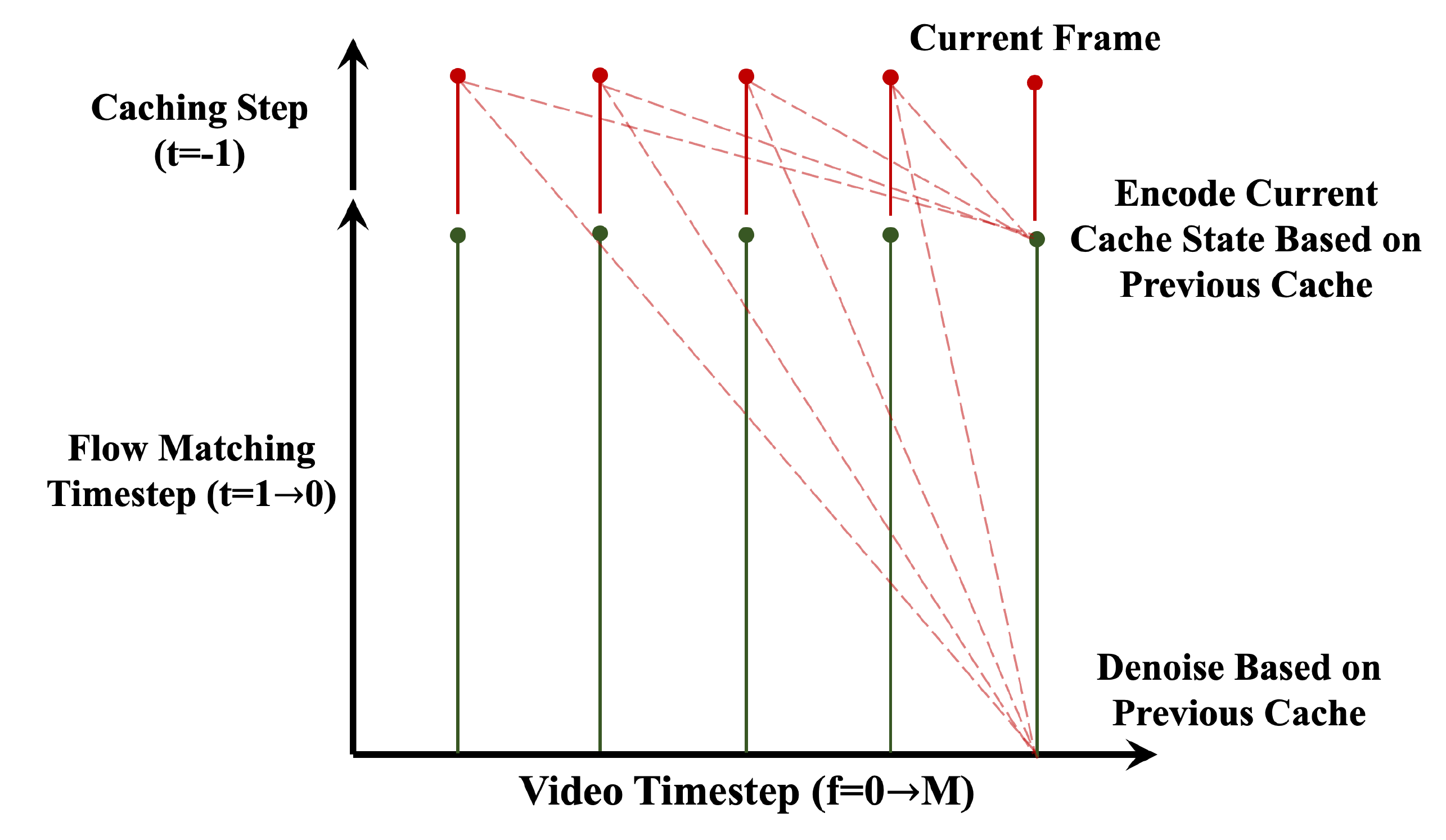}
    \vspace{-.24in}
    \caption{\textbf{KV Cache for Short-Video Modeling in FAR.} We additionally add a caching step to encode current decoded frame into the KV cache for autoregressive generation.}
    \label{fig:kv_cache}
    \vspace{-.1in}
\end{figure}

\begin{figure*}[!tb]
    \centering
    \includegraphics[width=\linewidth]{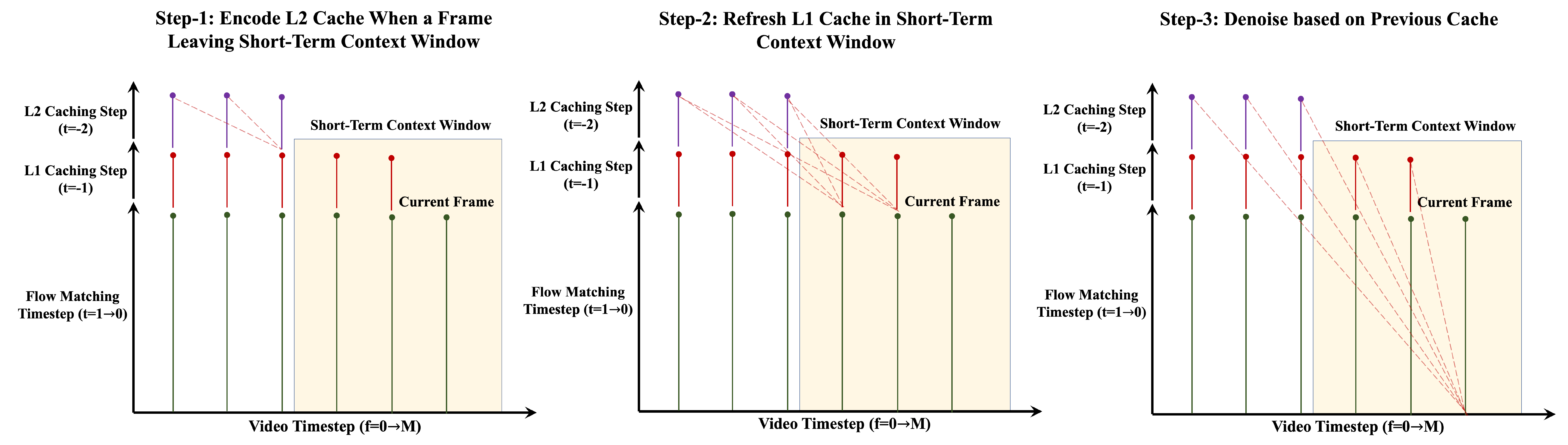}
    \vspace{-.28in}
    \caption{\textbf{Multi-Level KV Cache for Long-Context Video Modeling in FAR.} When a frame leaves the short-term context window, we encode it to the L2 cache and re-encode the L1 cache in the window. We then use those encoded KV cache for decoding the current frame. Note that we divide the process into three steps for better illustration, though it can be merged into a single forward pass in implementation.}
    \label{fig:multi_level_kv}
    \vspace{-.1in}
\end{figure*}

\subsection{Long-Context Video Modeling}
\label{sec:long_context}

\myPara{Test-Time Extrapolation \textit{vs.} Long-Video Training.}
Two promising approaches to achieving long-context modeling in language modeling are test-time extrapolation~\cite{blocntkaware,peng2023yarn,su2024roformer} and long-sequence fine-tuning~\cite{chen2023extending,chen2023longlora}. In video modeling, most efforts~\cite{zhao2025riflex,wang2023gen,lu2024freelong} have focused on test-time extrapolation to generate long videos. However, we question \textbf{\textit{whether test-time extrapolation can effectively solve long-context video modeling}}. To investigate this, we train FAR on action-conditioned short-video prediction and extend it to long-video prediction using various extrapolation techniques. As summarized in \tabref{tab:comp_tte}, test-time extrapolation results in significantly lower quality than the baseline sliding window approach, leading to poor predictions. Therefore, direct training on long videos may be necessary to achieve effective long-context video modeling. 

\myPara{Explosive Token Growth in Long Video Training.} Visual data contains redundancy, causing vision tokens to expand much faster than language tokens as context increases. For example, a video sequence of 128 frames requires more than 8K tokens, as illustrated in \figref{fig:asymetric_context}(a). Consequently, training on long videos becomes computationally prohibitive, as shown in \figref{fig:asymetric_context}(b, c).

\myPara{Our Solution: Long Short-Term Context Modeling.} To address explosive token growth in long-video training, we leverage the concept of \textbf{\textit{context redundancy}}. Specifically, in video autoregression, the current frame depends more heavily on nearby frames to capture local motion consistency, whereas earlier frames primarily serve as context memory. To eliminate context redundancy, we propose a long short-term context modeling with \textbf{\textit{asymmetric patchify kernels}}. As shown in \figref{fig:pipeline}(b), we maintain a high-resolution short-term context window to capture fine-grained temporal consistency, and a low-resolution long-term context window, where we apply a large patchify kernel to compress context tokens.
During training, given that the data has a maximum sequence length of $m$ frames, we fix the short-term context window to $n$ frames and randomly sample the long-term context frames from the range $[0, m-n]$.
The attention mask with long short-term context modeling is shown in \figref{fig:attention_mask}(b), where the long-term context uses fewer tokens. As demonstrated in \figref{fig:asymetric_context}(a), this strategy ensures that increasing the vision context length maintains a manageable token context length. With long short-term context modeling, we can reduce the cost and memory usage of the long-video training significantly, as shown in \figref{fig:asymetric_context}(b, c).
To prevent interference between long-term and short-term contexts, we adopt separate projection layers for each context, inspired by MM-DiT~\cite{esser2024scaling}. This approach results in a slightly larger number of parameters, referred to as FAR-Long in \tabref{tab:model_cfg}.

\subsection{Inference-Time KV Cache}
\label{sec:kv_cache}
\myPara{KV Cache for Short-Video Modeling.}
Due to the autoregressive nature of FAR, we can leverage KV-Cache to accelerate inference. As illustrated in \figref{fig:kv_cache}, for each frame, we first use the flow-matching schedule to decode it into the clean latent frame. We then introduce an additional caching step to encode the clean latent frame into the KV cache. As discussed in \secref{sec:basictrain}, we use timestep $t=-1$ to denote the clean context frame in the caching step. These KV caches are subsequently used for autoregressive decoding of future frames.

\myPara{Multi-Level KV Cache for Long-Video Modeling.}
In long-context video modeling, we employ long short-term context to reduce redundant visual tokens. To accommodate this, we introduce a multi-level KV cache. As illustrated in \figref{fig:multi_level_kv},
the frames in long-term context window is encoded into L2 cache (4 tokens per frame), while the frames in short-term context window is encoded into L1 cache. 
When decoding current frame but exceed the short-term context window, the earliest frame in the short-term context window is moved to the long-term context window and encode it into the L2 cache. Since this modifies the cache state, we subsequently re-encode the L1 cache of the frames in the short-term context window. The encoded cache is then used to decode the current frame. Note that in practice, these three steps can be merged into a single forward pass for efficiency.

\section{Experiment}

\subsection{Implementation Details}
We follow the DiT's structure~\cite{peebles2023scalable} to implement FAR. To compress video latents, we train a series of image DC-AE~\cite{chen2024deep} on the corresponding dataset, resulting in 64 tokens per frame. All models are trained from scratch without image pretraining.
We provide training hyperparameters and evaluation setting in the \tabref{tab:eval_setting}.

\begin{table}[!tb]
    \caption{\textbf{Quantitative Comparison of Conditional and Unconditional Video Generation on UCF-101.} We follow the evaluation setup of Latte~\cite{ma2024latte}. $\dag$ denotes FVD reported on 10,000 videos.}
    \label{tab:ucf101_results}
    \vspace{-.12in}
    \centering
    \resizebox{\linewidth}{!}{
    \begin{tabular}{l|ccc|cc}
    \toprule
    \multirow{2}{*}{\textbf{Methods}} & \multirow{2}{*}{\textbf{Type}} & \multirow{2}{*}{\textbf{Params}} & \textbf{Double} & \textbf{Cond. Gen} & \textbf{Uncond. Gen} \\
    &&&\textbf{Train Cost}& \textbf{FVD$_{2048}$} $\downarrow$ & \textbf{FVD$_{2048}$} $\downarrow$\\
    \midrule\midrule
    \multicolumn{6}{c}{\textbf{Resolution-128$\times$128}}\\
    \midrule\midrule
    MAGVITv2-MLM~\cite{yu2023language} & Non-AR & 307 M & \xmark & 58$\dag$ & - \\
    MAGVITv2-AR~\cite{yu2023language}  & Token-AR & 840 M & \xmark & 109$\dag$ & - \\
    TATS~\cite{ge2022long}        & Token-AR & 331 M & \xmark & 332 & 420 \\
    \midrule
    FAR-L (Ours) & Frame-AR & 457 M & \xmark & \textbf{99 (57$\dag$)} & \textbf{280} \\
    \midrule\midrule
    \multicolumn{6}{c}{\textbf{Resolution-256$\times$256}}\\
    \midrule\midrule
    LVDM~\cite{he2022latent}         & Video-DiT   & 437 M & \xmark & -   & 372 \\
    Latte~\cite{ma2024latte}        & Video-DiT   & 674 M & \xmark & -   & 478 \\
    CogVideo~\cite{hong2022cogvideo}     & Token-AR & 9.4 B     & \xmark & 626 & - \\
    OmniTokenizer~\cite{wang2024omnitokenizer}     & Token-AR & 650 M     & \xmark & 191 & - \\
    ACDIT~\cite{hu2024acdit}       & Frame-AR & 677 M & \cmark & 111 & - \\
    MAGI~\cite{zhou2025taming}         & Frame-AR & 850 M & \cmark & -   & 421 \\
    \midrule
    FAR-L (Ours)    & Frame-AR & 457 M & \xmark & 113 & 303 \\
    FAR-XL (Ours)    & Frame-AR & 674 M & \xmark & \textbf{108} & \textbf{279} \\
    \bottomrule
\end{tabular}
}
\vspace{-.2in}
\end{table}

\begin{table*}[!tb]
    \caption{\textbf{Quantitative Comparison on Short Video Prediction.} We follow the evaluation setup of MCVD~\cite{voleti2022mcvd} and ExtDM~\cite{zhang2024extdm}, where $c$ denotes the number of context frames and $p$ denotes the number of predicted frames.}
    \label{tab:video_prediction}
    \vspace{-.12in}
    \centering
    \begin{subtable}[t]{0.40\linewidth}
    \resizebox{\linewidth}{!}{
    \begin{tabular}{l|c|cccc}
        \toprule
        \multirow{2}{*}{\textbf{Methods}} & \multirow{2}{*}{\textbf{Params}} & \multicolumn{4}{c}{$c=4,\  p=12$} \\
        \cmidrule{3-6}
         & & \textbf{SSIM$\uparrow$} & \textbf{PSNR$\uparrow$} & \textbf{LPIPS$\downarrow$} & \textbf{FVD$\downarrow$} \\
        \midrule
        RaMViD~\cite{hoppe2022diffusion} & 235 M & 0.639 & 21.37 & 0.090 & 396.7 \\
        LFDM~\cite{ni2023conditional} & 108 M & 0.627 & 20.92 & 0.098 & 698.2 \\
        MCVD-cp~\cite{voleti2022mcvd} & 565 M & 0.658 & 21.82 & 0.088 & 468.1 \\
        ExtDM-K2~\cite{zhang2024extdm} & 119 M & 0.754 & 23.89 & 0.056 & 394.1 \\\midrule
        FAR-B (Ours) & 130 M & \textbf{0.818} & \textbf{25.64} & \textbf{0.037} & \textbf{194.1} \\
        \bottomrule
    \end{tabular}}
    \caption{\textbf{Evaluation on UCF-101 (64$\times$64)}}
    \end{subtable}\hfill
    \begin{subtable}[t]{0.58\linewidth}
    \resizebox{\linewidth}{!}{
    \begin{tabular}{l|c|cccc|cccc}
        \toprule
        \multirow{2}{*}{\textbf{Methods}} & \multirow{2}{*}{\textbf{Params}} & \multicolumn{4}{c|}{$c=2,\  p=14$} & \multicolumn{4}{c}{$c=2,\  p=28$} \\
        \cmidrule{3-10}
         & & \textbf{SSIM$\uparrow$} & \textbf{PSNR$\uparrow$} & \textbf{LPIPS$\downarrow$} & \textbf{FVD$\downarrow$} & \textbf{SSIM$\uparrow$} & \textbf{PSNR$\uparrow$} & \textbf{LPIPS$\downarrow$} & \textbf{FVD$\downarrow$} \\
        \midrule
        RaMViD~\cite{hoppe2022diffusion} & 235 M & 0.758 & 17.55 & 0.085 & 166.5 & 0.691 & 16.51 & 0.109 & 238.7 \\
        LFDM~\cite{ni2023conditional} & 108 M & 0.770 & 17.45 & 0.084 & 167.6 & 0.730 & 16.68 & 0.106 & 276.8 \\
        VIDM~\cite{mei2023vidm} & 194 M & 0.763 & 16.97 & 0.080 & 131.7 & 0.728 & 16.20 & 0.096 & 194.6\\
        MCVD-cp~\cite{voleti2022mcvd} & 565 M & 0.838 & 19.10 & 0.075 & 87.8 & 0.797 & 17.70 & 0.078 & 119.0\\
        ExtDM-K4~\cite{zhang2024extdm} & 121 M & 0.845 & 20.04 & 0.053 & \textbf{81.6} & 0.814 & 18.74 & 0.069 & \textbf{102.8}\\\midrule
        FAR-B (Ours) & 130 M & \textbf{0.849} & \textbf{20.87} & \textbf{0.038} & 99.3 & \textbf{0.819} & \textbf{19.40} & \textbf{0.049} & 144.3 \\
        \bottomrule
    \end{tabular}}
    \caption{\textbf{Evaluation on BAIR (64$\times$64)}}
    \end{subtable}
\end{table*}

\begin{table*}[!tb]
    \caption{\textbf{Quantitative Comparison on Long-Context Video Prediction.} We follow the evaluation setup of TECO~\cite{yan2023temporally}, where $c$ denotes the number of context frames and $p$ denotes the number of predicted frames.}
    \label{tab:long_context_prediction}
    \vspace{-.12in}
    \centering
    \begin{subtable}[t]{0.5\linewidth}
    \resizebox{\linewidth}{!}{
    \begin{tabular}{l|c|ccc|c}
        \toprule
        \multirow{2}{*}{\textbf{Methods}} & \multirow{2}{*}{\textbf{Params}} & \multicolumn{3}{c|}{$c=144,\  p=156$} & $c=36,\  p=264$ \\
        \cmidrule{3-6}
         & & \textbf{SSIM$\uparrow$} & \textbf{PSNR$\uparrow$} & \textbf{LPIPS$\downarrow$} & \textbf{FVD$\downarrow$} \\
        \midrule
        FitVid~\cite{babaeizadeh2021fitvid} & 165 M & 0.356 & 12.0 & 0.491 & 176 \\
        CW-VAE~\cite{saxena2021clockwork} & 111 M & 0.372 & 12.6 & 0.465 & 125 \\
        Perceiver AR~\cite{hawthorne2022general} & 30 M & 0.304 & 11.2 & 0.487 & 96 \\
        Latent FDM~\cite{harvey2022flexible} & 31 M & 0.588 & 17.8 & 0.222 & 181 \\
        TECO~\cite{yan2023temporally} & 169 M & \textbf{0.703} & 21.9 & 0.157 & \textbf{48} \\\midrule
        FAR-B-Long (Ours) & 150 M & 0.687 & \textbf{22.3} & \textbf{0.104} & 64 \\
        \bottomrule
    \end{tabular}}
    \caption{\textbf{Evaluation on DMLab (64$\times$64)}}
    \end{subtable}\hfill
    \begin{subtable}[t]{0.5\linewidth}
    \resizebox{\linewidth}{!}{
    \begin{tabular}{l|c|ccc|c}
        \toprule
        \multirow{2}{*}{\textbf{Methods}} & \multirow{2}{*}{\textbf{Params}} & \multicolumn{3}{c|}{$c=144,\  p=156$} & $c=36,\  p=264$ \\
        \cmidrule{3-6}
         & & \textbf{SSIM$\uparrow$} & \textbf{PSNR$\uparrow$} & \textbf{LPIPS$\downarrow$} & \textbf{FVD$\downarrow$} \\
        \midrule
        FitVid~\cite{babaeizadeh2021fitvid} & 176 M & 0.343 & 13.0 & 0.519 & 956 \\
        CW-VAE~\cite{saxena2021clockwork} & 140 M & 0.338 & 13.4 &  0.441 & 397 \\
        Perceiver AR~\cite{hawthorne2022general} & 166 M & 0.323 & 13.2 & 0.441 & 76 \\
        Latent FDM~\cite{harvey2022flexible} & 33 M & 0.349 & 13.4 & 0.429 & 167 \\
        TECO~\cite{yan2023temporally} & 274 M & 0.381 & 15.4 & 0.340 & 116 \\\midrule
        FAR-M-Long (Ours) & 280 M & \textbf{0.448} & \textbf{16.9} & \textbf{0.251} & \textbf{39} \\
        \bottomrule
    \end{tabular}}
    \caption{\textbf{Evaluation on Minecraft (128$\times$128)}}
    \end{subtable}
\end{table*}

\subsection{Quantitative Comparison}

\subsubsection{Video Generation}
\myPara{Dataset and Evaluation Settings.} We benchmark both unconditional and conditional video generation on the UCF-101 dataset~\cite{soomro2012ucf101}, which consists of approximately 13,000 videos. Following Latte~\cite{ma2024latte}, we use the entire dataset for training. 
For evaluation, we randomly sample 2,048 videos to compute FVD~\cite{unterthiner2019fvd} against the ground-truth videos. For conditional video generation, we set the guidance scale to 2.0 during inference.

\myPara{Main Results.}
From the results listed in \tabref{tab:ucf101_results}, we achieve state-of-the-art performance in both unconditional and conditional video generation. Specifically, Latte~\cite{ma2024latte} is based on video diffusion transformer, while OmniTokenizer~\cite{wang2024omnitokenizer} is based on Token AR. Our method significantly outperforms both. 
Furthermore, compared to recent frame-autoregressive models~\cite{hu2024acdit,zhou2025taming}, which require twice the training cost, FAR achieves superior performance without any additional training cost.

\subsubsection{Short-Video Prediction}
\myPara{Dataset and Evaluation Settings.}
We evaluate FAR on the UCF-101~\cite{soomro2012ucf101} and BAIR~\cite{ebert2017self} datasets, following the evaluation settings in MCVD~\cite{voleti2022mcvd} and ExtDM~\cite{zhang2024extdm}. 
We randomly sample 256 videos based on provided context frames, each with 100 different trajectories, and select the best trajectory to compute pixel-wise metrics. For FVD, we report the average over all trajectories.

\myPara{Main Results.} We summarize the results in \tabref{tab:video_prediction}. Unlike previous works such as MCVD~\cite{voleti2022mcvd} and ExtDM~\cite{zhang2024extdm}, which introduce complex multi-scale fusion strategies and optical flow, FAR achieves superior results on both datasets without requiring additional design.

\subsubsection{Long-Video Prediction}
\myPara{Dataset and Evaluation Settings.}
We benchmark long-context video modeling results on action-conditioned video prediction using the Minecraft and DMLab datasets~\cite{yan2023temporally}. The Minecraft dataset contains approximately 200K videos, while the DMLab dataset contains about 40K videos. Each video consists of 300 frames with action annotations. We follow the evaluation setup in TECO~\cite{yan2023temporally}, which uses 144 observed context frames to predict 156 future frames and compute pixel metrics. Additionally, we compute FVD on 264 generated frames based on 36 context frames.

\myPara{Main Results.} We summarize the results in \tabref{tab:long_context_prediction}. The previous work, TECO~\cite{yan2023temporally}, adopts aggressive downscaling for all frames to reduce tokens for temporal modeling, creating a trade-off between training efficiency and prediction accuracy. In contrast, FAR employs long short-term context modeling, effectively achieving the lowest prediction error (\ie, LPIPS) without prohibitive computation cost.

\subsection{Qualitative Comparison}
\begin{figure*}[!tb]
    \centering
    \includegraphics[width=\linewidth]{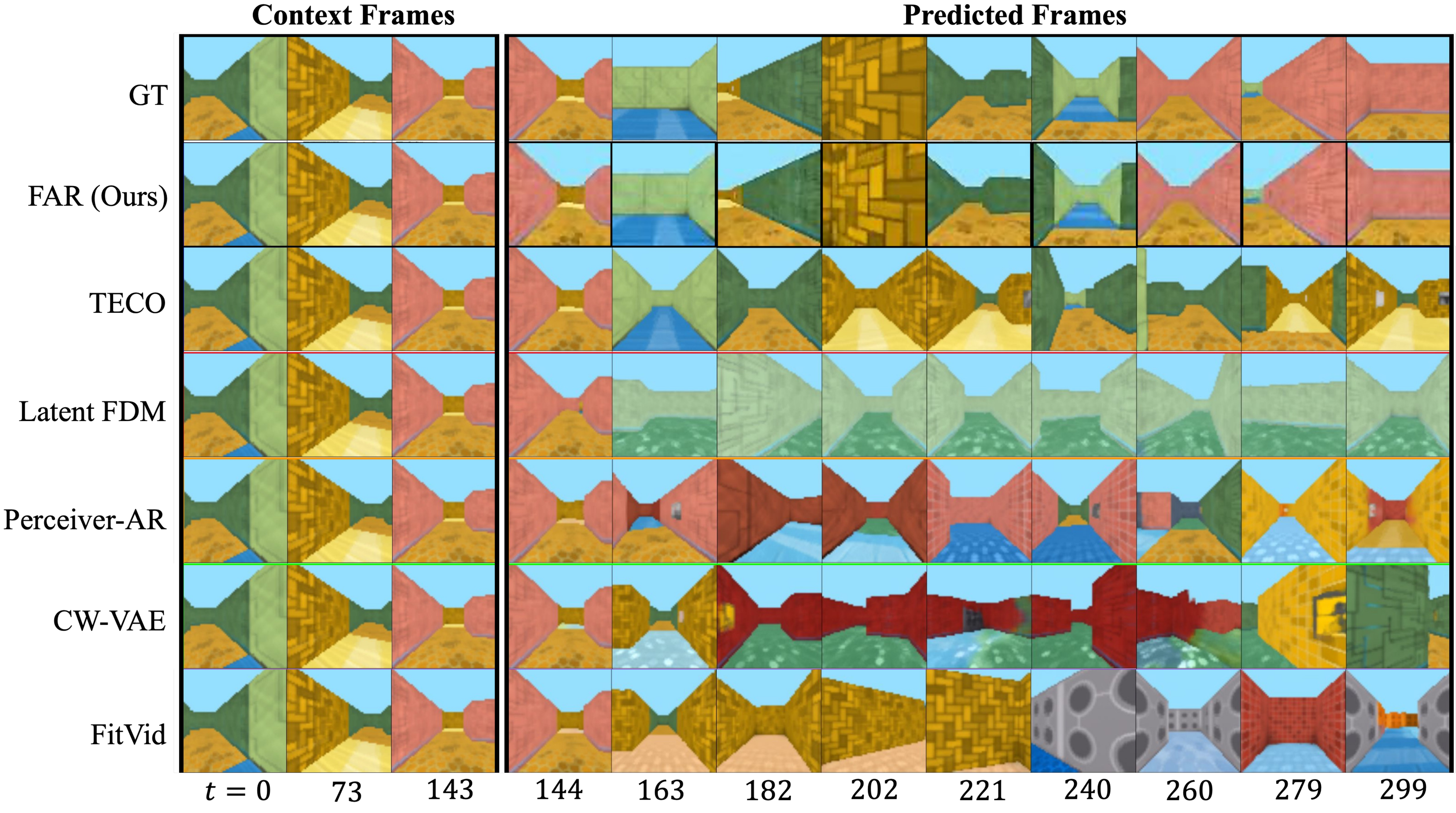}
    \vspace{-.25in}
    \caption{\textbf{Qualitative Comparison of Long-Context Video Prediction on DMLab.} FAR fully utilizes the long-range context (144 frames), resulting in more consistent prediction (156 frames) compared to previous methods.}
    \label{fig:dmlab_sample1}
    \vspace{-.1in}
\end{figure*}

We present a qualitative comparison of long-video prediction in \figref{fig:dmlab_sample1}. Compared to previous methods, FAR effectively utilizes the observed context and generates predictions that most closely resemble the ground truth, demonstrating its ability to leverage long-range context.

\subsection{Ablation Study}

\myPara{How to decide patchify kernel for remote context?}
The key to selecting an appropriate patchify kernel lies in whether the compressed latent representation can reliably retain information from past observations. Since patchifying is a spatial-to-channel transformation, we hypothesize that it should satisfy the condition $c \times c \times d \leq D$, where $c$ is the patchify kernel size, $d$ is the latent dimension, and $D$ is the model’s channel dimension.
For example, in our experiment, the latent dimension is 32 and the model dimension is 768. Using a patchify kernel of size $4 \times 4$, we get $4 \times 4 \times 32 = 512 < 768$, which suggests that nearly all input information can be preserved during patchification.
As demonstrated in \tabref{tab:ablation_kernel}, the $[4,4]$ patchify kernel significantly reduces training cost, enabling feasible training on long videos, without sacrificing prediction accuracy compared to a larger $[8,8]$ kernel.

\begin{figure}[!tb]
    \centering
\includegraphics[width=\linewidth]{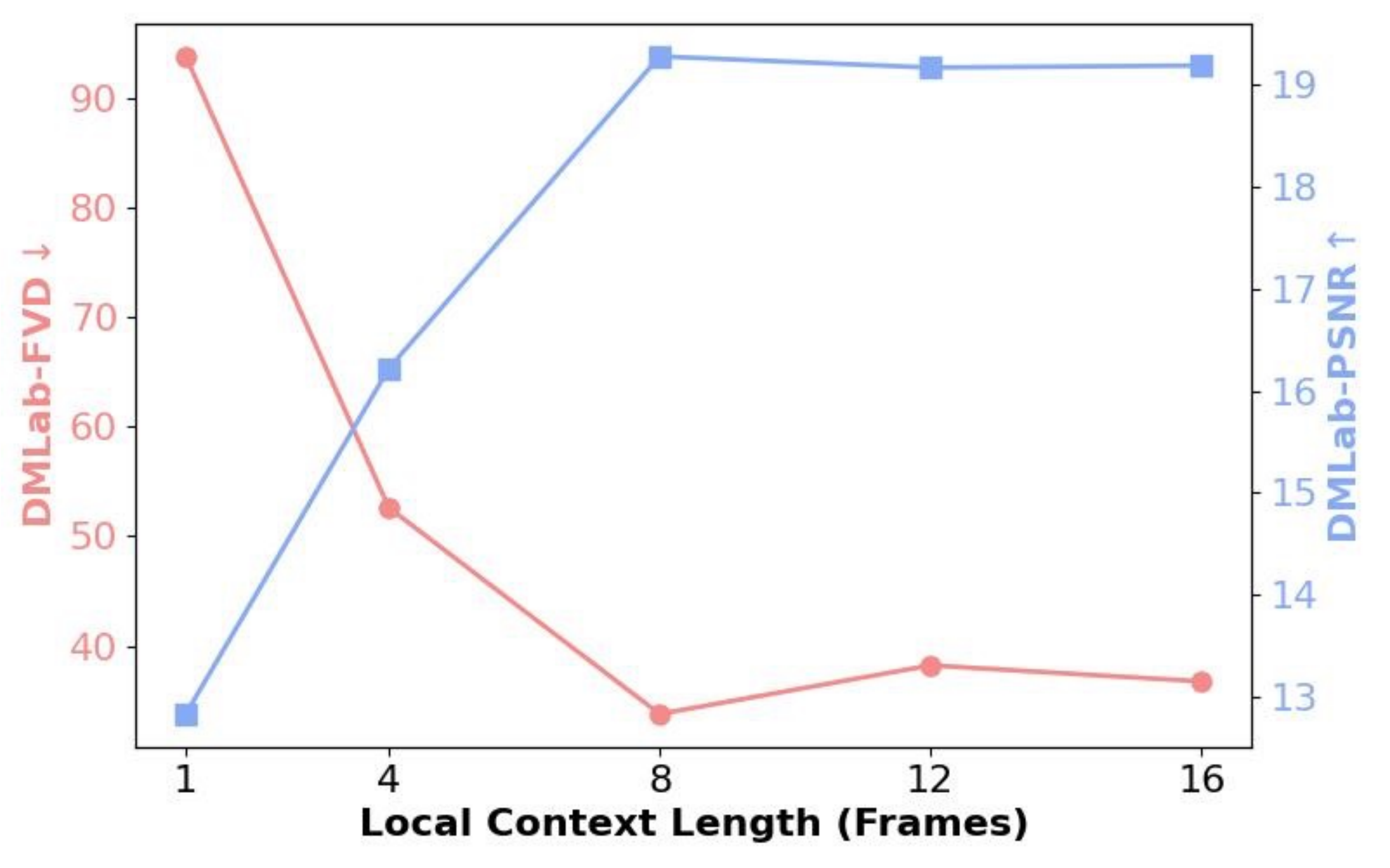}
    \vspace{-.25in}
    \caption{\textbf{Ablation Study of the Short-Term Context Window Size.} Performance saturates as the window size increases.}
    \label{fig:windowsize}
    \vspace{-.1in}
\end{figure}

\myPara{How to decide the local context length?}
We gradually increase the local context length during training and observe that the performance converges at certain short-term context length (\ie, 8 frames in \figref{fig:windowsize}). Further increasing the local context length significantly raises the training cost without improving performance. This experiment verifies the presence of context redundancy in video autoregressive modeling. Therefore, we select the saturate point as the optimal short-term context length.

\begin{table}[!tb]
    \caption{\textbf{Ablation Study of Stochastic Clean Context (SCC) on UCF-101.} Stochastic clean context mitigates the training-inference discrepancy in observed context, leading to improved performance.}
    \label{tab:ablation_cleanctx}
    \vspace{-.13in}
    \centering
    \resizebox{\linewidth}{!}{\begin{tabular}{l|cccc}
        \toprule
        \multirow{2}{*}{\textbf{Methods}}  
 & \multicolumn{4}{c}{$c=1,\  p=15$} \\
        \cmidrule{2-5}
        & \textbf{SSIM$\uparrow$} & \textbf{PSNR$\uparrow$} & \textbf{LPIPS$\downarrow$} & \textbf{FVD$\downarrow$} \\
        \midrule
        w/o. SCC & 0.540 & 16.42 & 0.211 & 399 \\
        w/. SCC & \textbf{0.596} & \textbf{18.46} & \textbf{0.187} & \textbf{347} \\
        \bottomrule
    \end{tabular}}
\end{table}

\myPara{Effect of Stochastic Clean Context.}
We have visualized the effectiveness of stochastic clean context in \figref{fig:stochastic_clean_context}. Based on the quantitative evaluation of video prediction in \tabref{tab:ablation_cleanctx}, FAR with stochastic clean context achieves significantly improved performance.

\begin{figure}[!tb]
    \centering
\includegraphics[width=\linewidth]{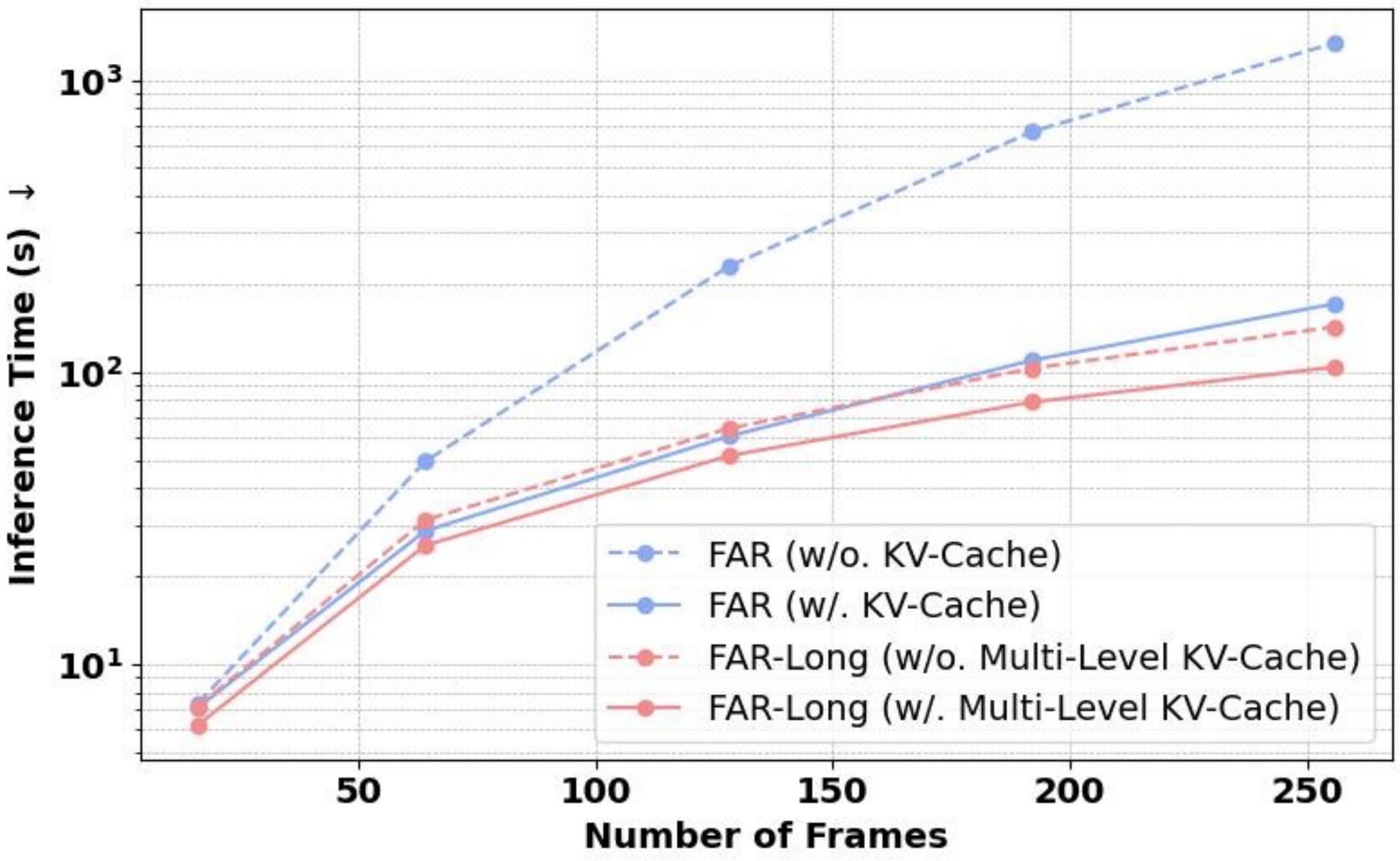}
    \vspace{-.25in}
    \caption{\textbf{Ablation Study of the KV Cache.} FAR-Long with proposed multi-level KV cache achieves the best speedup on long videos.}
    \label{fig:ablation_kvcache}
    \vspace{-.1in}
\end{figure}

\myPara{Effect of the KV Cache on Inference Speedup.}
As shown in ~\figref{fig:ablation_kvcache}, the baseline FAR model, which samples without using long short-term context or KV cache requires approximately 1341 seconds to generate a 256-frame video. When the KV cache is introduced, the inference time is significantly reduced to 171 seconds. Finally, by incorporating both long short-term context and the corresponding multi-level KV cache, the sampling time further decreases to approximately 104 seconds. These results demonstrate that KV caching, especially when combined with long short-term context modeling, significantly improves sampling efficiency for long video generation.

\begin{table}[!tb]
    \centering
    \caption{\textbf{Ablation Study on the Patchify Kernel of Distant Context.} Larger patchify kernels significantly reduce training cost.}
    \label{tab:ablation_kernel}
    \resizebox{\linewidth}{!}{
        \begin{tabular}{c|ccccc}
            \toprule
            \makecell{\textbf{Patchify}\\\textbf{Kernel}} & 
            \textbf{SSIM$\uparrow$} & 
            \textbf{PSNR$\uparrow$} &
            \textbf{LPIPS$\downarrow$} &
            \textbf{FVD$\downarrow$} & 
            \makecell{\textbf{Training}\\\textbf{Memory}} \\ 
            \midrule
            $[1, 1]$ & - & - & - & - & OOM \\
            $[2, 2]$ & 0.570 & 19.1  & 0.156 & 38  & 38.9 G \\
            $[4, 4]$ & 0.576 & 19.3 & 0.153 & 34 & 15.3 G \\
            $[8, 8]$ & 0.558 & 18.6  & 0.171  & 33 & 0.9 G \\
            \bottomrule
        \end{tabular}
    }
\end{table}

\section{Conclusion}

In this paper, we systematically investigate long-context video modeling using the proposed Frame Autoregressive Model (FAR). First, We show that direct test-time extrapolation is insufficient for effective long-context video modeling, highlighting the necessity of efficient long-video training. We then identify \textbf{\textit{context redundancy}} as a key bottleneck in video autoregression. To address this, we propose a long short-term context modeling strategy with \textbf{\textit{asymmetric patchify kernels}}, a simple yet effective method to eliminate redundant context and significantly reduce the training cost for long videos. Extensive experiments validate the effectiveness of FAR in handling long-context video generation and highlight a promising direction for the evolution of next-generation video generative models—shifting the focus from short-term temporal consistency to long-term world modeling.

\myPara{Limitations.}
The primary limitation lies in the lack of scaled-up experiments. Although FAR demonstrates great potential, we still lack large-scale training on text-to-video generation datasets. Additionally, restricted by the available datasets, we only experiment with FAR on up to 300 frames (about 20 seconds), not fully investigating its ability on minute-level videos.

\myPara{Future Work.}
One future direction is to scale up FAR and benchmark it against video diffusion transformers on large-scale text-to-video generation tasks. Additionally, we plan to simulate a longer video dataset (on the minute level) to better evaluate the model’s long-context capabilities. Finally, it would be interesting to explore whether FAR’s long-context modeling can enable video-level in-context learning.

{
\bibliographystyle{IEEEtran}
\bibliography{reference}
}

\begin{table*}[!tb]
\caption{
\textbf{Experimental Configurations of FAR.} We follow the evaluation settings from Latte~\cite{ma2024latte}, MCVD~\cite{zhang2024extdm}, and TECO~\cite{yan2023temporally}.}
\vspace{-.12in}
\label{tab:eval_setting}
\centering
\resizebox{\linewidth}{!}{
\begin{tabular}{lccccccc}
\toprule
\multirow{2}{*}{\textbf{Hyperparameters}} & \multicolumn{2}{c}{\textbf{Short-Video Generation}} & \multicolumn{2}{c}{\textbf{Short-Video Prediction}} & \multicolumn{2}{c}{\textbf{Long-Video Prediction}} \\
& \textbf{Cond. UCF-101} & \textbf{Uncond. UCF-101} & \textbf{BAIR} & \textbf{UCF-101} & \textbf{Minecraft} & \textbf{DMLab} \\ \midrule\midrule
\multicolumn{7}{c}{\textbf{Dataset Configuration}}\\\midrule\midrule
Resolution & 256/128 & 256/128 & 64 & 64 & 128 & 64 \\
Total Training Samples & 13,320 & 13,320 & 43,264 & 9,624 & 194,051 & 39,375 \\
\midrule\midrule
\multicolumn{7}{c}{\textbf{Training Configuration}}\\\midrule\midrule
Training Cost (H100 Days) & 12.7 & 12.7 & 2.6 & 3.6 & 18.2 & 17.5 \\
Batch Size & 32 & 32 & 32 & 32 & 32 & 32 \\
Latent Size & 8$\times$8 (DC-AE~\cite{chen2024deep}) & 8$\times$8 (DC-AE~\cite{chen2024deep}) & 8$\times$8 (DC-AE~\cite{chen2024deep}) & 8$\times$8 (DC-AE~\cite{chen2024deep}) & 8$\times$8 (DC-AE~\cite{chen2024deep}) & 8$\times$8 (DC-AE~\cite{chen2024deep}) \\
Training Sequence Length & 16 & 16 & 32 & 16 & 300 & 300 \\
LR & $1\times 10^{-4}$ & $1\times 10^{-4}$ & $1\times 10^{-4}$ & $1\times 10^{-4}$ & $1\times 10^{-4}$ & $1\times 10^{-4}$ \\
LR Schedule & constant & constant & constant & constant & constant & constant \\
Warmup Steps & - & - & - & - & 10K & 10K \\
Total Training Steps & 400K & 400K & 200K & 200K & 1M & 1M \\
Stochastic Clean Context & 0.1 & 0.1 & 0.1 & 0.1 & 0.1 & 0.1 \\
Short-Term Context Window & 16 & 16 & 32 & 16 & 16 & 16 \\
Patchify Kernel for Distant Context & - & - & - & - & [4, 4] & [4, 4]\\
\midrule\midrule
\multicolumn{7}{c}{\textbf{Evaluation Configuration}}\\\midrule\midrule
Samples & 4$\times$2048 & 4$\times$2048 & 100$\times$256 & 100$\times$256 & 4$\times$256 & 4$\times$256 \\
Guidance Scale & 2.0 & - & - & - & 1.5 & 1.5 \\
Reference Work & Latte~\cite{ma2024latte} & Latte~\cite{ma2024latte} &  MCVD~\cite{zhang2024extdm} & MCVD~\cite{zhang2024extdm} & TECO~\cite{yan2023temporally} & TECO~\cite{yan2023temporally} \\
\bottomrule
\end{tabular}}
\end{table*}

\section{Appendix}

\subsection{Experimental Settings}
As shown in \tabref{tab:eval_setting}, we list the detailed training and evaluation configurations of FAR. For the ablation study in this paper, we halve the training iterations while keeping other settings the same.

\subsection{Qualitative Comparison}
\begin{figure*}[!tb]
    \centering
    \includegraphics[width=\linewidth]{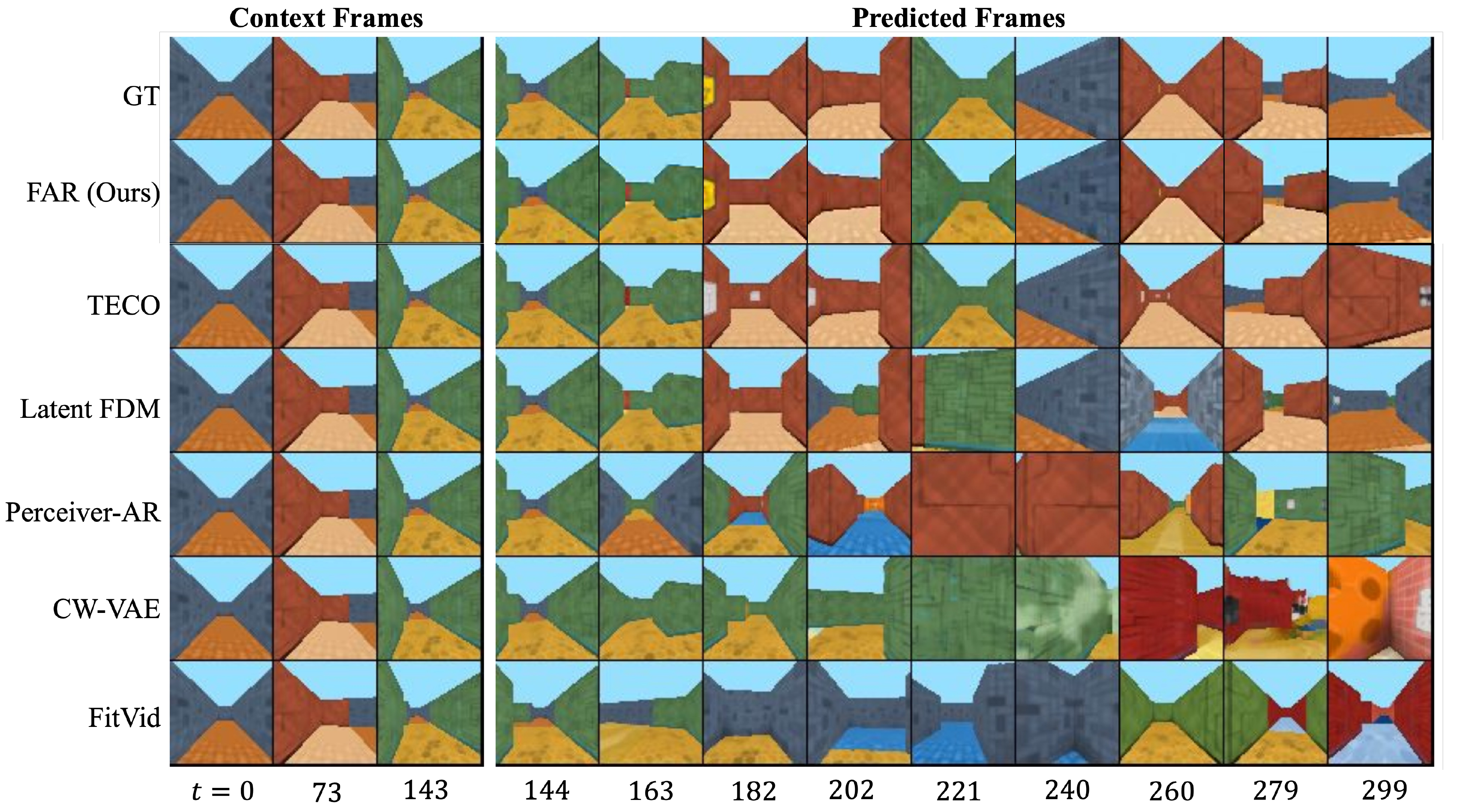}
    \vspace{-.25in}
    \caption{\textbf{Qualitative Comparison of Long-Context Video Prediction on DMLab.} FAR fully utilizes the long-range context (144 frames), resulting in more consistent prediction (156 frames) compared to previous methods.}
    \label{fig:dmlab_sample2}
\end{figure*}

\begin{figure*}[!tb]
    \centering
    \includegraphics[width=\linewidth]{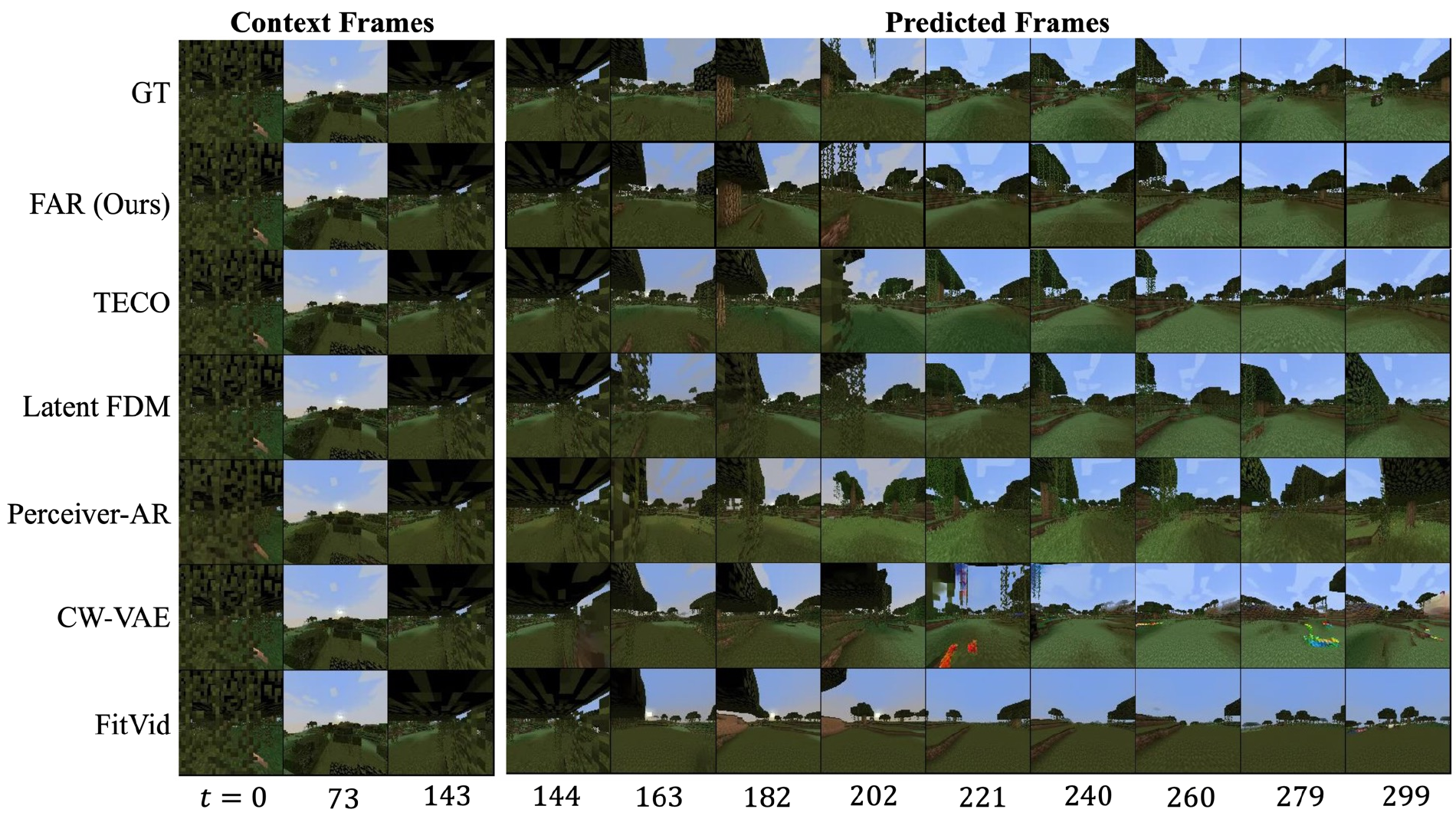}
    \vspace{-.25in}
    \caption{\textbf{Qualitative Comparison of Long-Context Video Prediction on Minecraft.} FAR fully utilizes the long-range context (144 frames), resulting in more consistent prediction (156 frames) compared to previous methods.}
    \label{fig:minecraft_sample1}
\end{figure*}

We provide additional visualization of long-video prediction results on DMLab and Minecraft in \figref{fig:dmlab_sample2} and \figref{fig:minecraft_sample1}. From the results, FAR better exploits the provided context and provides more consistent results in later predictions compared to previous works.

\end{document}